# GraphFlow: An Architecture for Formally Verifiable Visual Workflows Enabling Reliable Agentic AI Automation

Drewry H. Morris V Luis Valles, MBA Reza Hosseini Ghomi, MD, MSE

*MedFlow, Inc.*

**Abstract**

GraphFlow is a visual workflow system designed to improve the reliability of agentic AI automation in multi-step, mission-critical processes. In these workflows, small errors compound rapidly: under an idealized model of independent steps, a ten-step process with 90% per-step reliability completes successfully only 35% of the time. Existing workflow platforms deliver durable execution and observability but offer few semantic correctness guarantees, and agentic systems plan at inference time, making behavior sensitive to prompt variation and difficult to audit. GraphFlow is designed to close this gap by treating workflow diagrams as the executable specification, a single artifact defining data scope, execution semantics, and monitoring. At compile time, a restricted class of diagrams is specified to produce reusable automations whose contracts (preconditions, postconditions, and composition obligations) are required to be proof-checked before admission to a shared library. At runtime, a durable engine records outcomes in an append-only event log and can enforce contracts at system boundaries, supporting replay, retries, and audit. Swimlanes make trust boundaries explicit, separating verified logic from external systems, human judgment, and AI decisions. A year-long pilot across three clinical sites executed 8,728 cohort-enrolled workflow runs with a 97.08% completion rate under an early prototype without the verified-core subsystem (observed failures were localized to external integrations). The formal semantics and proof-checked admission described here are specified and under active development. Evaluation of the verified core is reserved for a follow-up paper.

## 1. Introduction

Agentic AI workflows suffer from high error rates in multi-step processes, where errors compound across steps. Under an idealized model in which step failures are independent and per-step reliability is identical, a 10-step workflow with 90% per-step reliability has approximately 35% overall reliability (since $0.9^{10} \approx 0.35$), meaning roughly 65% of workflows would fail. To reach 90% overall reliability under the same model, each step must be about 99% reliable (since $0.99^{10} \approx 0.90$). These assumptions rarely hold exactly in practice: real workflows exhibit correlated failures (shared upstream outages) and mitigating behavior (retries, human fallback), so end-to-end rates can be worse or better than the iid bound predicts. The qualitative point stands: small per-step error rates accumulate rapidly across long workflows, making traditional approaches insufficient for mission-critical applications. GraphFlow is designed to improve agentic reliability by shifting planning into diagrams and contracts, then executing those diagrams under durable semantics with auditable traces.

This problem is acute across mission-critical domains. High-profile deployments in healthcare have repeatedly demonstrated that AI systems which perform well on benchmarks can fail once integrated with real clinical workflows: IBM Watson for Oncology failed to reach meaningful clinical deployment after substantial investment, with MD Anderson terminating its flagship partnership over integration problems and misalignment with real clinical data [1], [2]. More broadly, Raji et al. document that many deployed AI systems simply do not work outside controlled environments, developing a taxonomy of functionality failures across case studies and arguing that the burden of proof for functionality should rest with deployers rather than users [3]. In medical AI specifically, trust depends on specialized expertise for safe development, verification, and operation. Failures in deployed systems risk eroding public confidence in both the technology and the institutions that adopt it [4]. Recent evaluations of large language models reinforce this concern: on the Cancer-Myth benchmark, which tests patient questions containing false presuppositions, LLMs frequently fail to correct dangerous misconceptions, making them unreliable for unsupervised clinical use [5]. Healthcare chatbots in particular require careful attention to human-AI interaction design and transparency, not just model accuracy [6]. The same classes of failure, legacy system integration, data quality gaps, and misaligned assumptions between automated and human steps, recur across customer service, manufacturing, and logistics deployments [3], and have surfaced publicly in incidents such as airline chatbot misinformation that produced legal liability when a tribunal held the carrier responsible for its chatbot's incorrect refund statements [7].

These failures demonstrate that agentic AI workflows require additional verification, monitoring, and integration capabilities to operate safely in mission-critical deployments where accuracy is paramount and errors can have severe consequences. A fundamental issue is that many agentic workflows are non-deterministic: given the same inputs, they may produce different outputs due to AI model variability, prompting inconsistencies, or stochastic decision-making. This non-determinism makes verification, debugging, and reliability assurance extremely difficult. Determinism improves reproducibility and debugging, but it does not by itself imply semantic correctness or domain validity. Traditional low-code platforms validate workflows at runtime, creating additional risks when deploying AI-generated workflows in critical domains.

For agentic AI, the reliability gap is often a planning gap. Agents must infer multi-step processes from natural language, choose tools, sequence actions, and recover from failures, making behavior sensitive to prompt variation, model stochasticity, and implicit assumptions. In GraphFlow, agent actions are constrained to authoring, selecting, parameterizing, and launching pre-approved workflows, while execution is performed by the workflow runtime. This shifts agentic work from ad hoc tool-call planning to workflow authoring and invocation: the agent proposes or selects a diagram, binds a cohort and parameters, and launches execution under the diagram's semantics.

GraphFlow's architecture bridges human expertise and AI-driven automation through three interconnected pillars. The core insight is that humans and AI have complementary strengths: humans excel at strategic judgment, contextual understanding, and noticing when something is wrong at a system level, while modern models excel at pattern-heavy transformation, synthesis, and anomaly detection in large datasets [8], [9], [10]. In practice, many human-in-the-loop systems still fail because they encourage automation bias, where people over-trust machine output and disengage from critical review [11], [12]. GraphFlow is designed to keep humans in the loop at the right layer: humans validate intent, scope, and process structure by reviewing diagrams, contracts, and boundaries, while the runtime executes those diagrams deterministically and produces auditable traces.

GraphFlow is grounded in systems thinking: reliability failures often originate at boundaries between tools, teams, and handoffs when responsibilities and assumptions are implicit. By making responsibility explicit via swimlanes (runtime, external systems, AI and humans), the workflow becomes reviewable as an end-to-end system rather than a collection of isolated steps [13].

## 1.1. Two Execution Modes at a Glance

GraphFlow supports two execution modes, distinguished throughout this paper:

- **Verified core (compile-time).** A restricted class of diagrams that are acyclic and sequential, with verified nodes confined to lanes intended for static reasoning. These compile into reusable automations whose contracts (preconditions, postconditions, composition obligations) must be proof-checked before the automation is admitted to the shared library.
- **Durable runtime.** Full production diagrams may include long-running behavior, retries, waits, and controlled cycles (for example, a retry loop from a decision back to an earlier step). These execute under a durable engine that records nondeterministic outcomes in an append-only event log, supporting deterministic replay and optional runtime contract enforcement at boundaries.

The verified core is a subset of what the runtime can execute: every verified-core automation is also runnable under the durable engine, but only acyclic, sequential diagrams are eligible for verified-core admission. Examples in this paper are labeled accordingly so readers can tell at a glance which mode they illustrate. Full semantics for both modes are given in Appendix D.

## 1.2. Related Work

GraphFlow builds on substantial prior work in workflow orchestration, durable execution, formal methods, and process modeling. This section surveys the landscape the paper draws from, then states what GraphFlow adds to it.

### 1.2.1. Prior Art

**Durable execution engines.** Systems such as Temporal [14] treat workflows as code running against an append-only event log, so a workflow survives restarts, retries, and long waits by replaying deterministically from the log. Durable execution yields reproducibility and operational robustness, but does not by itself establish semantic correctness for a given domain.

**Workflow reference models and patterns.** The Workflow Management Coalition reference model [15] and the workflow patterns catalog [16] defined a control-flow vocabulary (sequence, parallel split, deferred choice, exception handling) that recurs across modern engines. These remain a useful taxonomy for what a workflow runtime must support.

**Cloud state machines and DAG schedulers.** AWS Step Functions [17] provides a managed state-machine service with built-in retries, timeouts, and error handling. Apache Airflow [18] popularized the directed acyclic graph (DAG) pattern for scheduled data pipelines, with strong testing and monitoring ecosystems. Both target operational execution rather than formal verification.

**BPMN and enterprise process modeling.** BPMN 2.0 [19] and engines such as Camunda [20] provide a standardized graphical notation for business processes, widely used in regulated industries because diagrams are readable by domain experts and engineers alike. BPMN emphasizes governance and auditability, but does not produce machine-checkable correctness proofs.

**Contracts and proof-carrying code.** Hoare logic [21] introduced preconditions, postconditions, and composition reasoning as the basis for proving program correctness. Proof-carrying code [22] extended this pattern to mobile code, attaching machine-checkable proofs to compiled artifacts so a consumer can verify them without trusting the producer. GraphFlow applies the same pattern to its automations: each one carries proof obligations that must be discharged before admission to a shared library.

**Agentic AI reliability.** Recent work highlights the compounding error rate of multi-step agent workflows [23]. Studies of human-AI interaction document that automation bias and over-trust erode human-in-the-loop safeguards when reviewers disengage [11], [12].

#### 1.2.2. Positioning

We do not claim novelty in durable execution, retries, timers, human-in-the-loop steps, or visual workflow modeling. GraphFlow's contribution is treating diagrams as executable specifications that generate both reviewable artifacts **and** machine-checkable proof obligations, making workflows simultaneously operational and verifiable.

Concretely, GraphFlow contributes:

- **Diagram-as-specification**: a single source of truth for workflow structure, execution semantics, and monitoring, reducing diagram-code drift.
- **Two modes of correctness guarantees**: compile-time verification for reusable automations, and runtime enforcement through durable execution with deterministic replay.
- **Generated artifacts**: reviewable, versioned code and tests suitable for standard engineering workflows, enabling code review and integration with existing development practices.
- **Proof artifacts**: machine-checkable proof obligations for each automation, validated by a proof assistant, ensuring correctness claims are backed by formal verification.
- **Contract-first composition**: `requires`/`ensures` and composition checking as first-class workflow constructs, enabling safe reuse and composition of verified automations.
- **Agentic workflow grounding**: workflows are first-class executable artifacts that agents select and execute, reducing prompt-level variability by moving process structure into diagrams, contracts, and durable semantics.

GraphFlow differs from existing workflow systems primarily in what is treated as authoritative, what is verified, and how boundary assumptions are modeled. It shifts the comparison axis in three ways:

- **Proof-carrying workflow automations** [22]: a restricted subset of diagrams compiles into reusable workflow automations with explicit `requires`/`ensures`, generated code and tests, and machine-checkable proof obligations that must be satisfied before admission to the automation library.
- **Contract-checked composition**: composition is treated as a compile-time correctness problem, where postconditions must imply downstream preconditions, enabling reuse with preserved assumptions rather than informal handoffs.
- **Explicit boundary modeling**: swimlanes act as verification-relevant metadata, marking which nodes are eligible for compile-time reasoning and which interactions are external or human assumptions subject to runtime enforcement and audit. Formally, swimlanes are represented by a labeling function $\lambda : V \to L$ from diagram nodes to lanes (see Appendix D.3).

**Comparison Table**

| System | Execution | Verification |
| --- | --- | --- |
| Temporal [14] | Durable replay | Reproducibility + durability |
| AWS Step Functions [17] | Managed state machine | Operational guarantees |
| Apache Airflow [18] | Scheduled DAG runs | Testing/monitoring culture |
| BPMN engines [19], [20] | Token-flow/state | Governance + auditability |
| **GraphFlow (this paper)** | **Verified generation + durable runtime** | **Formal specifications + formal verification** |

Deterministic replay and durability provide reproducibility and strong operational behavior. Formal verification provides semantic guarantees relative to a specification (contracts, semantics, and proof obligations). GraphFlow treats these as complementary: determinism supports debugging and auditability, while verification supports correctness claims.

## 1.3. Three Pillars of GraphFlow

GraphFlow's architecture addresses reliability through three interconnected pillars, each preventing specific failure modes in agentic AI workflows.

### 1.3.1. Cohort Search

Cohort search fixes the execution boundary by selecting an auditable data scope. The data lake architecture enables graph-based search to identify precise cohorts matching workflow criteria using inclusion and exclusion criteria (entities with/without specific tags). This tag-based approach reduces the risk of executing on the wrong dataset by making cohort selection explicit, auditable, and reviewable. For agentic AI, this shifts data-scope selection from implicit prompting to a reusable definition that can be reviewed and monitored.

Intuitively, a cohort is just "the set of resources matching a query", for example, "all sales reports scheduled but not yet completed in FY 2025." The formalism below names the pieces so we can reason about them precisely.

A cohort $\Theta$ is the result of a query over the universe of tenant resources $\Xi$. A query $Q \in \mathbb{Q}$ is a pair $(\rho, \Phi)$ selecting resources of a given resource type $\rho$ (drawn from the set of types $\Sigma_\Xi$) that match a set of filters $\Phi$, and evaluates to a cohort $\Theta \subseteq \Xi$ via the evaluation function $q$:

$$Q = (\rho, \Phi), \quad q : Q \to \Theta, \quad \Theta \subseteq \Xi$$

Below is an example written in GraphFlow Language (GFL), our text-based language for defining core constructs:

```
query "Sales Reports Pending":
  description: |
    Pending sales reports for FY 2025

  resource-type: :report
  ext-type: "SalesReport"

  filters:
    - with: :scheduled
    - without: :completed
    - field: :date1
      operator: :gte
```

```
      value: "2025-01-01"
    - field: :date1
      operator: :lt
      value: "2026-01-01"
```

### 1.3.2. Formal Automation

Formal automation executes workflow diagrams under a formal semantics, compiling them into reusable automations with verified contracts and checked composition. Deterministic replay supports reproducibility, but correctness derives from contracts and verification.

Intuitively, a GraphFlow diagram is a directed graph whose nodes are workflow steps (tasks, decisions, external calls, etc.) and whose edges are sequential or control-flow transitions, annotated with metadata such as swimlane assignment, node type, and action bindings. A compiler converts diagrams that meet the structural restrictions (acyclic, sequential, verified nodes in verifier-eligible lanes) into reusable automations with machine-checked contracts. The notation below names these pieces.

A compiler $\mathcal{C}$ maps a compilable diagram $D$ (drawn from the admissible set $\mathbb{D}_c$) to an automation $A$. Each diagram consists of a directed graph $G = (V, E)$ with nodes $V$ and edges $E$, together with a tuple of metadata assignments $\mu$ for swimlane, node type, layout, and action bindings. See Appendix D.3 for the formal definition:

$$A = \mathcal{C}(D), \quad D \in \mathbb{D}_c, \quad D = (G, \mu), \quad G = (V, E)$$

The executable artifacts in the GraphFlow runtime are the diagrams $\mathbb{D}$, integrations $\mathbb{I}$, and automations $\mathbb{A}$:

$$\mathbb{X} = \mathbb{D} \cup \mathbb{I} \cup \mathbb{A}$$

A trigger $K = (\odot, \Omega)$ pairs an executable artifact $\odot \in \mathbb{X}$ with a binding $\Omega$ whose source cohort is $\Theta$ and whose call target is a diagram $D$ (that is, $\Omega_{\text{source}} = \Theta$ and $\Omega_{\text{calls}} = D$). The trigger invokes its target for each member $\xi$ of the cohort:

$$\forall \xi \in \Theta : \text{calls}\left(\odot\right)$$

Below is an example of a **runtime-mode** diagram written in GFL. Note the back-edge from the decision on line 4 (`:no--> 1`) that routes rejected reports back to the start. Diagrams with such cycles execute under the durable runtime but are not eligible for compile-time verified-core admission, which requires acyclicity (see “Verification Model”). A smaller acyclic example suitable for verification is provided in Appendix A.

```
diagram [blueprint] "Sales Report Submission Process":
  description: |
    Process for submitting sales reports

  swimlanes:
    - "Sales"
    - "COO"
    - "Accounting"

  model:
    1. [task] "Fill out Sales report" @sales --> 2:
      action:
        calls: (:submit-report)
    2. [report] "Sales Report" @sales --> 3:
```

```
    ext-type: "SalesReport"
  3. [meeting] "Review Report" @accounting --> 4:
    assigned: [:coo, :sales]
    action:
      calls: (:schedule-meeting)
  4. [decision] "Approve?" @coo :yes--> 5 :no--> 1:
    action:
      calls: (:request-approval {
        .yes: (:eq $.submitted :approved)
      })
  5. [task] "Submit to accounting" @accounting:
    action:
      calls: (:send-email)
```

A complete GFL example of a **verified-core-eligible** (acyclic, sequential) diagram with contracts and mathematical properties is provided in Appendix A.

#### 1.3.3. Operational Dashboards

Operational dashboards validate and monitor results through visualization, providing human-in-the-loop verification and response. Dashboards monitor diagram execution status across cohorts and use the same cohort search capabilities for monitoring, creating a closed-loop system where cohort identification, execution, and monitoring share the same underlying data model.

Metrics aggregate query results over time:

$$M = (Q, \oplus), \quad \oplus \in \Sigma_M$$

where $M$ computes $\oplus$ on a scheduled basis and $\Sigma_M = \{\text{count}, \text{sum}, \text{avg}\}$.

Below is an example written in GFL:

```
metric "Sales Reports Count":
  description: |
    Count of pending Sales reports over time

  query: :sales-reports-pending
  aggregation: :count
```

The diagram-as-specification model unifies the three pillars through shared notation:

$$K \overset{\text{source}}{\to} Q \overset{\text{cohort}}{\to} \Theta \overset{\text{calls}}{\to} D \overset{\text{monitor}}{\to} M$$

A trigger $K$ has a source query $Q$ to identify a cohort $\Theta$, to call diagram $D$ for each $\xi \in \Theta$, and metrics $M$ monitor the same cohort via the same query infrastructure.

### 1.4. System Architecture

GraphFlow separates specification, verification, execution, and user interaction into distinct components to keep compile-time assurance and runtime durability independent.

Terminology:

- **formal specification** = GFL semantics + contracts
- **formal verification** = proof-checked obligations for library admission
- **durable execution** = event-log replay + operational enforcement

## 1.5. Verification Model: Compile-Time and Runtime Guarantees

GraphFlow separates compile-time verification from runtime enforcement. Formal automation supplies verified components that are reused by cohort search and operational dashboards. The guarantees are defined by what is verified at compile time versus what is enforced at runtime:

**Verified-core automations (compile-time verified)**: A restricted subset of diagrams that are acyclic and sequential and whose verified nodes live in lanes intended for static reasoning. These compile into reusable automations with explicit `requires`/`ensures`, generated code and tests, and proof obligations that must be accepted before admission to the automation library. Each admitted automation is formally verified with respect to its declared contracts under the formal semantics and stated action assumptions (Appendix D).

**Durable runtime workflows (runtime enforced)**: Full production diagrams may include long-running behavior, retries, waits, and controlled cycles. These execute under a durable engine that records nondeterministic outcomes in an append-only event log and can enforce contracts at boundaries when enabled. GraphFlow guarantees deterministic replay when all nondeterministic outcomes used by workflow logic are recorded in the event log. Replay reads from the log rather than consulting nondeterministic sources.

The boundary between these modes is explicit in the diagram via swimlanes: nodes that cross into `"external"`, `"human"`, or `"AI"` lanes are treated as effect boundaries whose outcomes are assumptions recorded for replay and optionally validated by runtime guards. External APIs, human decisions, and AI decisions are modeled as assumptions through explicit contracts and, where available, idempotency protocols. We refer to these as the two **modes** of correctness guarantees throughout the paper: a compile-time verified core and runtime-enforced guarantees. For agentic automation, these two modes distinguish between reusable, specification-backed capabilities that an agent may safely invoke (verified-core automations) and workflows that require explicit monitoring and governance due to interaction with effectful systems and human judgment.

### 1.5.1. Guarantees

GraphFlow's guarantees are semantic guarantees relative to explicit contracts and formal semantics, not correctness-by-replay.

Guarantees (scoped):

- **Compile-time (library admission):** contract soundness for the workflow-automation semantics (Appendix D). Composition compatibility (postconditions imply downstream preconditions). Structural admissibility (e.g., rejects cycles). Verified-core admission requires verified lane assignment, structural restrictions, and accepted proof obligations.
- **Runtime (durable execution):** deterministic replay for fixed initial state + event log (Appendix D). Retry safety under stated idempotency assumptions. Optional boundary guards for contracts and invariants.

Non-goals: determinism improves auditability but does not imply domain truth. External systems, AI and human judgments remain assumptions.

## 1.6. Human-in-the-Loop and Boundary Modeling

GraphFlow makes human responsibility explicit rather than implicit. Swimlanes classify each node as runtime, external system, AI or human, which determines (1) what can be verified, (2) what must be recorded for replay, and (3) where approvals or checkpoints are required. This keeps judgment at the right layer: humans validate intent and approve boundary actions, while the runtime executes the diagram deterministically and produces auditable traces. Escalation is a first-class behavior (pause/approve/route), so failures or contract violations become reviewable events rather than silent drift.

## 1.7. Design

The following sections detail the core technical components and execution model.

### 1.7.1. Agent Loop

We use "agent" in the standard sense of an iterative decision process that observes a system, maintains internal state, and chooses actions. Let $s_t$ denote the environment state at step $t$. The agent receives an observation $o_t = \mathrm{Obs}(s_t)$ and maintains an internal memory state $m_t$. Given $(o_t, m_t)$, the agent selects an action $a_t$ according to a policy $\pi$:

$$a_t = \pi(o_t, m_t)$$

After executing $a_t$, the environment transitions to a new state $s_{t+1}$ and the agent updates memory:

$$m_{t+1} = \mathrm{U}(m_t, o_t, a_t)$$

In GraphFlow, the agent's actions operate over workflow artifacts, not over arbitrary tool-call sequences. At each turn, the agent may (1) retrieve relevant blueprints from the repository, (2) propose a new diagram or a bounded edit, (3) bind a cohort and parameters, and (4) launch a run (one-off) or deploy a scheduled/background automation. Execution of nodes is performed by the GraphFlow runtime under the selected diagram. The agent is not called to decide each next step unless an explicit AI node is part of the diagram.

$$a_t \in \{\mathrm{retrieve}, \mathrm{author}, \mathrm{edit}, \mathrm{bind}, \mathrm{launch}, \mathrm{deploy}\}$$

Control flow is therefore defined by the selected diagram, while nondeterminism is confined to explicit boundary nodes (external systems, AI and human judgment) whose outcomes are recorded for audit and replay.

### 1.7.2. Agent modes: authoring vs invocation

GraphFlow uses the same agent loop in two modes. In **authoring mode**, the agent behaves like a consultant: it helps draft or refine diagrams, recommends pre-approved automations, and helps define the cohort and dashboard checks needed for safe operation. In **invocation mode**, the agent resolves a natural-language request to an existing approved diagram, binds parameters, and launches a one-off run. In both modes, the agent's freedom is constrained to repository-backed workflows and bounded edits. The runtime engine executes the workflow.

### 1.7.3. Agent Integration: Process Selection vs Process Synthesis

GraphFlow is designed to serve as a verified execution framework for agentic systems operating in integration-heavy domains. The key shift is that agents do not synthesize workflows ad hoc. Instead, they select existing diagrams or propose bounded modifications, bind inputs, and execute under formal semantics. This reduces behavioral variance by stabilizing control flow, isolating nondeterminism at explicit boundaries, and producing replayable execution traces suitable for audit and human review.

### 1.7.4. Diagram-as-Specification Architecture

The visual diagram **is** the execution specification: generated artifacts and durable runtime behavior are mechanically derived from it.

This architecture is designed to reduce drift and improve reviewability:

- Generated artifacts are derived mechanically from the diagram.
- For long-running instances, the runtime executes the diagram semantics directly.
- Humans review diagrams as a first-class artifact rather than relying solely on AI-generated workflow structure.

#### 1.7.4.1. Canonical Artifact and Execution Modes

GraphFlow supports two execution paths, both driven from the same diagram definition. These execution paths are orthogonal to the correctness guarantee modes (compile-time verified vs runtime enforced) described in "Verification Model":

- **Code generation**: For diagrams that become verified automations, GraphFlow generates code (and proof obligations) that become the deployable execution artifact. The diagram remains the source of truth. Any change to behavior requires regenerating artifacts from the updated diagram.
- **Durable runtime execution**: For long-running instances (including runtime cycles, retries, and checkpointing), the runtime engine executes the diagram semantics using an event log. This mode prioritizes durability and observability over minimal runtime footprint.

#### 1.7.4.2. Deployment Modes

GraphFlow can be deployed in two primary ways:

- **Local deployment (small scale)**: run and iterate on diagrams locally as a developer tool for small-scale workflows and validation.
- **Cloud deployment (large scale)**: run GraphFlow as a multi-tenant service with durable execution, operational dashboards, and tenant/workspace isolation.

Diagrams are directed graphs $G = (V, E)$ equipped with assignment functions (Appendix D), such as the swimlane assignment $\lambda : V \to L$. Swimlanes are semantic classifications (e.g., runtime, external system, human) that determine verification eligibility and audit requirements. Nodes in verified lanes are interpreted by the verifier, while nodes in boundary lanes are treated as effectful steps whose outcomes are recorded for replay and may be validated by runtime guards.

Execution semantics are defined over the diagram's graph structure.

**Node types (runtime · compile-time):**

- **task** – work item · local computation
- **meeting** – coordination · external system call
- **report** – information artifact · in-memory value
- **object** – physical entity/material · persistent storage
- **decision** – decision point · conditional branch
- **queue** – ordered queue · iteration/loop
- **milestone** – phase marker · return/final output
- **wait** – delay/interrupt · delay/event listener
- **diagram** – run another diagram · module call

Nodes may contain **subdiagrams** (nested models) for decomposition/iteration. All nodes are assigned to **swimlanes** that determine verification eligibility and responsibility.

**Edge types:**

- **Sequential flow** – to
- **Control flow** – yes, no, maybe

Runtime determinism is provided by the event log (see "Verification Model").

Because the diagram is the authoritative execution specification, generated code and runtime behavior remain mechanically synchronized with the visual model. This preserves reviewability over time and allows both humans and automated tooling to reason about workflow behavior using the same artifact (see Appendix C for a visual example).

#### 1.7.5. Scalability and Execution

Verification cost scales with diagram size and predicate complexity. Proof checking dominates when enabled. Runtime cost is dominated by external I/O and event-log growth (roughly proportional to executed steps). Practical limits are set by CI timeouts and memory budgets. Large deployments require backpressure (concurrency and rate limits) and observability. The runtime provides append-only event logs, checkpointing, retries, and deterministic replay, skipping previously executed effects via idempotency keys.

### 1.8. Agentic Authoring, Invocation, and Trust Boundary

In an agentic setting, GraphFlow constrains autonomy into a reviewable workflow model. Instead of producing arbitrary tool-call sequences, an agent selects an existing workflow diagram (or proposes a bounded edit), supplies parameters such as cohort inputs and thresholds, and executes it under the diagram's semantics. Nondeterminism is confined to explicit boundary nodes representing external systems, human judgment, or explicit AI decision nodes. Control flow remains stable, auditable, and replayable.

For one-off requests (for example, "Send a proposal to customer X"), the agent matches the request to an approved diagram, fills required parameters, and launches a single run. The resulting execution is durable, audited, and replayable, with the same operational guarantees as scheduled automations.

GraphFlow treats AI as an untrusted assistant: it may propose diagrams and contracts, but admission to the automation library requires mechanical checks (and proofs when enabled). The boundary between the compile-time verified core and runtime-enforced guarantees is defined in "Verification Model."

#### 1.8.1. What is guaranteed (and what is not)

GraphFlow's strongest claims are semantic guarantees relative to an explicit specification (contracts + formal semantics), not correctness-by-replay.

The scope of guarantees is defined in "Verification Model" and formalized in Appendix D. Compile-time guarantees apply only to verified-core automations admitted to the library. Runtime guarantees apply to durable execution with deterministic replay and optional boundary guards. External systems, AI and human decisions remain assumptions recorded for audit and replay.

#### 1.8.2. Formal model and AI authoring

The formal semantics and theorem statements that scope the above claims are given in Appendix D. The paper's high-level guarantee statements should be read as "correctness relative to the stated semantics and assumptions," not as claims about the correctness of external systems, AI or human decisions.

- AI may propose: diagram structure, `requires`/`ensures`, invariants, boundary guards, and refactorings.
- The system must validate mechanically: parsing and typing, verified-core admissibility, contract and composition obligations, and (when enabled) proof-assistant-checked obligations for admitted automations.

- Humans remain responsible for: intent validation, domain correctness of contracts, and operational choices about enabling runtime checks, approvals, and deployment, so incorrect proposals are rejected at compile time or fail as explicit contract violations.

## 1.9. Testing and Operational Assurance

- Verification establishes properties relative to the formal semantics and declared contracts for the verified core.
- Property-based testing exercises the executable semantics over large input spaces, complementing proofs for practical coverage and regression resistance [24].
- Monitoring and dashboards validate distributional expectations, data freshness, and runtime invariants on live cohorts, supporting human-in-the-loop oversight without treating human judgment as a provable step.

For mission-critical deployments, this combination is practical: proofs constrain the verified core, replay and logs make execution reproducible, and monitoring catches violations of real-world assumptions.

## 1.10. Failure Modes and Limitations

GraphFlow improves reliability, but mission-critical deployments must account for failure modes outside workflow semantics.

- Bad specifications: incorrect or incomplete contracts can yield “verified” behavior that is wrong for the domain.
- Data quality and staleness: incorrect or stale data can drive correct execution over incorrect facts.
- External integration behavior: partial failures, inconsistent third-party semantics, and non-idempotent endpoints can violate assumptions.
- Human decision errors: approvals can be wrong. Escalation and review policies must be designed to manage this risk.

GraphFlow’s design mitigates (but does not eliminate) these risks via explicit boundary modeling, audit trails, runtime guards, and operational controls.

## 1.11. Security and Trust Model

GraphFlow is designed with a conservative posture for mission-critical automation. The architecture targets alignment with NIST SP 800-53 (Moderate baseline) control families as a design reference [25] (not a certification claim).

Trust boundaries:

- User/UI: untrusted input. Changes are authenticated, authorized, and audited.
- AI assistant: untrusted. Produces suggestions only.
- Generator + verifier: trusted build step. Emits artifacts only when checks (and proofs, when enabled) pass.
- Runtime engine: trusted to enforce isolation, durability, and audit logging.
- External systems: untrusted. Treated as assumptions and optionally guarded at runtime.

Core operational assumptions:

- Tenant/workspace isolation is enforced across storage and execution.
- Secrets are resolved at runtime in least-privilege contexts and are not embedded into generated artifacts.

- Generated artifacts are treated as supply-chain inputs: reviewed, versioned, and reproducibly regenerated.

## 1.12. Evaluation Plan

This paper focuses on architecture and verification scope. We evaluate GraphFlow using two complementary approaches: (1) empirical evidence from a production pilot (Empirical Evaluation) and (2) a forward measurement plan for the fully verified system (Evaluation Plan).

- **Prevented error classes**: contract violations become explicit failures. Composition mismatches are rejected before admission to the automation library. Verified-core restrictions prevent unsupported structures from being "verified by accident."
- **Cost model**: compile-time cost scales with diagram size and predicate complexity. Proof checking is the dominant cost when enabled. Runtime overhead is dominated by enabled guard evaluation and is often small relative to external I/O.
- **Scalability measurements**: compile-time checking time vs diagram size, proof-checking time vs obligation size, runtime throughput vs cohort size, event-log growth vs executed steps, replay performance for representative workflows.

## 1.13. Empirical Evaluation

This section reports observations from a year-long pilot deployment of an early GraphFlow prototype, evaluating the feasibility of the diagram-as-specification approach in real clinical workflows. The pilot demonstrates durable execution, explicit boundary modeling, and failure localization in practice. The deployed workflow is reconstructed in GFL in Appendix B with a corresponding visual diagram in Appendix C.

### 1.13.1. Deployment Context

The pilot ran from January 2025 through December 2025 across three anonymized clinical sites, each operating under different organizational and workflow conditions:

- **Clinic $\boldsymbol{\alpha}$** (telemedicine practice): active for the full study period (January-December 2025).
- **Clinic $\boldsymbol{\beta}$** (family medicine practice): onboarded in October 2025.
- **Clinic $\boldsymbol{\gamma}$** (single clinical location within a larger, multi-department healthcare organization): onboarded in October 2025.

All workflows integrated with a production electronic health record (EHR) system. The pilot used a cohort-enrollment model in which every patient enrolled in a clinic's active cohort was bound to exactly one workflow run (see "Cohort Search"). Across the deployment period, **8,728 unique patients** were enrolled, yielding **8,728 distinct workflow runs**. The one-to-one mapping between patients and runs is by design, not coincidence. Each reported run used the initial EHR order-creation call as its instrumented completion gate. Downstream clinical actions and conditional follow-up orders existed operationally but were not counted as additional completion gates in this metric.

### 1.13.2. Workload Summary

Across all sites and active periods, the system executed 8,728 workflow runs (8,473 completed, 255 errored), yielding an aggregate per-run completion rate of 97.08% (error rate 2.92%). We report three related but distinct rates in this evaluation:

- **Per-patient enrollment**: 8,728 patients, each producing one workflow run.
- **Per-run completion**: 97.08%, the fraction of workflow runs that reached a terminal completed state.

- **Per-boundary success**: separately, the instrumented EHR order-creation boundary completed successfully in 97.08% of attempts. Each reported run contained one such gating EHR order-creation attempt, so per-run completion and per-boundary success are numerically identical by construction. This should not be read as a claim that all downstream clinical or EHR interactions had the same success rate. Those steps were not fully instrumented as GraphFlow completion gates in the prototype.

Clinic $\alpha$ accounted for the majority of executions due to its longer deployment period. Clinics $\beta$ and $\gamma$ contributed fewer total executions because of later onboarding, though Clinic $\gamma$ exhibited higher per-period volume due to organizational scale and authorization complexity. Clinic $\gamma$'s lower success rate reflects the same authorization complexity in a multi-department organization, discussed further in "Failure Analysis."

Failures were recorded uniformly through the durable execution engine and represent executions that reached an errored state.

#### 1.13.3. Execution Summary

| **Clinic** | **Period** | **Completed** | **Errored** | **Success** |
|---|---|---|---|---|
| Clinic $\alpha$ | Jan–Dec | 6,914 | 73 | 98.96% |
| Clinic $\beta$ | Oct–Dec | 102 | 0 | 100.00% |
| Clinic $\gamma$ | Oct–Dec | 1,457 | 182 | 88.90% |
| **Total** | – | **8,473** | **255** | **97.08%** |

#### 1.13.4. Boundary Reliability and Multi-Step Failure

The introduction describes a compounding failure effect in multi-step workflows. The pilot data provides an empirical estimate of reliability for a representative external-system boundary: EHR order creation completed successfully in 97.08% of attempts. This empirically grounds the compounding argument introduced with the idealized $0.9^{10}$ example at a realistic observed per-boundary rate. Although this evaluation does not provide end-to-end node-level telemetry for every step in the care pathway, this boundary reliability is sufficient to illustrate how errors can accumulate across multi-step agentic workflows.

As an illustrative upper bound, assume a workflow requires **k** boundary calls whose failures are independent and identically distributed with per-call success probability **p**. Then expected end-to-end success scales as **p^k**. Under this assumption and the observed boundary success rate ($\mathbf{p} = 0.9708$), a workflow with five such calls would have an expected success rate of approximately 86%, and a workflow with ten calls approximately 74%. Real clinical workflows rarely satisfy strict independence and identical per-call reliability: failures at a given boundary are often correlated (shared upstream outages, per-clinic authorization gaps) or mitigated (retries, human fallback), so actual compounded rates can be worse or better than the iid model predicts. The qualitative point stands: small per-step error rates accumulate rapidly across long workflows, and localizing failures to explicit boundaries is a prerequisite for preventing invalid assumptions from silently propagating across steps.

#### 1.13.5. Failure Analysis

Observed failures were overwhelmingly attributable to external-system behavior rather than workflow control logic. Common causes included authorization failures (for example, providers lacking permission to place orders within specific departments) and upstream data integrity issues (such as duplicated or deleted patient records).

These failures occurred at explicit external-system boundaries and were recorded as such in the execution log. No failures were attributed to incorrect diagram structure or execution semantics.

#### 1.13.6. Interpretation and Limitations

This pilot reflects an early GraphFlow prototype that predated the formal specification and verification framework described in this paper. Workflow automations were generated from diagrams, but proof obligations, formal semantics, and verified composition were not yet implemented.

Accordingly, these results should be interpreted as evidence of:

1. The practicality of treating diagrams as executable specifications in real-world, multi-tenant clinical workflows.
2. The effectiveness of durable execution and event logging in localizing failures at system boundaries.
3. The tendency of reliability failures in agentic workflows to originate from external integrations and organizational constraints rather than internal control flow.

Several steps in the deployed care pathway (including clinician actions and background tagging jobs) were executed operationally but not recorded as GraphFlow node-level events. As a result, reported success and failure rates reflect only the instrumented external integration boundary rather than the full end-to-end clinical process. Nevertheless, these observations highlight where reliability failures actually surface in practice: at explicit system and organizational boundaries.

Future empirical evaluation will measure the impact of the verified core on end-to-end reliability, verification cost, and operational overhead once contract-checked generation and composition are deployed. Together, these results reinforce the central reliability argument of this paper: even modest per-boundary failure rates can significantly erode overall success in multi-step workflows, motivating the need for explicit specification, verification, and durable execution.

# 2. Status and Roadmap

This paper describes the GraphFlow architecture as it exists today: a specified design with a partially implemented runtime, a deployed clinical pilot, and a formal verification framework under active development. The sections below detail what is implemented, specified, and forthcoming.

## 2.1. Implemented today

- GraphFlow Language (GFL) for authoring diagrams, queries, metrics, and triggers.
- Durable runtime execution with an append-only event log, deterministic replay, retries, and idempotent execution.
- Cohort search and operational dashboards backed by the same query infrastructure.
- Swimlane-based boundary modeling at runtime, with outcomes at external, human, and AI boundaries recorded as audit-quality events.
- A year-long, multi-tenant clinical pilot whose results are reported in “Empirical Evaluation.”

## 2.2. Specified and in active development

- Formal semantics for the compile-time subset (Appendix D).
- `requires`/`ensures` contract checking and composition obligations along diagram edges.
- Proof-checked admission for verified-core automations.
- Generator that emits code and proof obligations from admissible diagrams.
- Proof-assistant integration for discharging obligations.

### 2.3. Forthcoming

- Production deployment of the verified core at pilot sites, planned for 2026.
- End-to-end empirical evaluation comparing verified-core automations against the unverified prototype measured in this paper.
- Scalability measurements of verification time, proof-checking cost, and runtime overhead as a function of diagram size and predicate complexity.
- A follow-up paper with these empirical results incorporated.

Accordingly, the correctness claims in this paper should be read as claims about the architecture and its formal semantics, not as empirical claims about a production-deployed verified core. The pilot numbers reported here reflect the pre-verification prototype and should be interpreted as evidence for the practicality of the diagram-as-specification and durable-execution layers, not as evidence for the verification layer, which is not yet in production.

## 3. Conclusion

GraphFlow improves the reliability of agentic automation by shifting planning from ad hoc tool-call sequences into a diagram-as-specification artifact with explicit scope, semantics, and assumptions. A formal specification (GFL semantics and contracts) enables verified generation of reusable automations, so agents primarily select and parameterize pre-approved processes from a repository rather than synthesizing multi-step plans on the fly.

At runtime, durable execution provides deterministic replay and auditable traces, while swimlanes make trust boundaries explicit by separating verified logic from external, human, and AI judgment. This keeps humans in the loop at the right points, localizes nondeterminism at clear boundary nodes, and turns reliability into an engineering discipline grounded in explicit specifications, checkable assumptions, and durable execution.

## 4. References

### 4.1. External References

## Bibliography

[1] C. Schmidt, "M. D. Anderson Breaks With IBM Watson, Raising Questions About Artificial Intelligence in Oncology," *JNCI: Journal of the National Cancer Institute*, vol. 109, no. 5, p. djx113, 2017, doi: 10.1093/jnci/djx113.

[2] E. Strickland, "How IBM Watson Overpromised and Underdelivered on AI Health Care." [Online]. Available: https://spectrum.ieee.org/how-ibm-watson-overpromised-and-underdelivered-on-ai-health-care

[3] I. D. Raji, I. E. Kumar, A. Horowitz, and A. D. Selbst, "The Fallacy of AI Functionality," *arXiv preprint arXiv:2206.09511*, 2022, [Online]. Available: https://arxiv.org/abs/2206.09511

[4] T. P. Quinn, M. Senadeera, S. Jacobs, S. Coghlan, and V. Le, "Trust and Medical AI: The challenges we face and the expertise needed to overcome them," *arXiv preprint arXiv:2008.07734*, 2020, [Online]. Available: https://arxiv.org/abs/2008.07734

[5] W. B. Zhu *et al.*, "Cancer-Myth: Evaluating Large Language Models on Patient Questions with False Presuppositions," *arXiv preprint arXiv:2504.11373*, 2025, [Online]. Available: https://arxiv.org/abs/2504.11373

[6] M. Jovanović, M. Baez, and F. Casati, "Chatbots as conversational healthcare services," *arXiv preprint arXiv:2011.03969*, 2020, [Online]. Available: https://arxiv.org/abs/2011.03969

[7] C. C. Rivers, "Moffatt v. Air Canada." [Online]. Available: https://canlii.ca/t/k2spq

[8] M. Mitchell, *Artificial Intelligence: A Guide for Thinking Humans*. Farrar, Straus, Giroux, 2019.

[9] F. Chollet, "On the Measure of Intelligence," *arXiv preprint arXiv:1911.01547*, 2019, [Online]. Available: https://arxiv.org/abs/1911.01547

[10] Apple Machine Learning Research, "The Illusion of Thinking: Understanding the Strengths and Limitations of Reasoning Models via the Lens of Problem Complexity." [Online]. Available: https://machinelearning.apple.com/research/illusion-of-thinking

[11] R. Parasuraman and V. Riley, "Humans and Automation: Use, Misuse, Disuse, Abuse," *Human Factors*, vol. 39, no. 2, pp. 230–253, 1997, doi: 10.1518/001872097778543886.

[12] S. Amershi *et al.*, "Guidelines for Human-AI Interaction," in *Proceedings of the 2019 CHI Conference on Human Factors in Computing Systems*, ACM, 2019, pp. 1–13. doi: 10.1145/3290605.3300233.

[13] W. E. Deming, *The New Economics for Industry, Government, Education*. MIT Press, 1993.

[14] "Temporal: Durable Execution." [Online]. Available: https://docs.temporal.io/temporal

[15] D. Hollingsworth, "Workflow Management Coalition: The Workflow Reference Model," technical report TC0–3, 1995. [Online]. Available: https://wfmc.org/wp-content/uploads/2022/09/tc003v11.pdf

[16] W. M. P. van der Aalst, A. H. M. ter Hofstede, B. Kiepuszewski, and A. P. Barros, "Workflow Patterns," *Distributed and Parallel Databases*, vol. 14, no. 1, pp. 5–51, 2003, doi: 10.1023/A:1022883727209.

[17] "AWS Step Functions." [Online]. Available: https://aws.amazon.com/step-functions/

[18] "Apache Airflow." [Online]. Available: https://airflow.apache.org/

[19] "Business Process Model and Notation (BPMN) Version 2.0." [Online]. Available: https://www.omg.org/spec/BPMN/2.0/

[20] "Camunda Platform." [Online]. Available: https://camunda.com/

[21] C. A. R. Hoare, "An Axiomatic Basis for Computer Programming," *Communications of the ACM*, vol. 12, no. 10, pp. 576–583, 1969, doi: 10.1145/363235.363259.

[22] G. C. Necula, "Proof-Carrying Code," in *Proceedings of the 24th ACM SIGPLAN-SIGACT Symposium on Principles of Programming Languages (POPL)*, ACM, 1997, pp. 106–119. doi: 10.1145/263699.263712.

[23] M. Kleppmann, "Prediction: AI will make formal verification go mainstream." [Online]. Available: https://martin.kleppmann.com/2025/12/08/ai-formal-verification.html

[24] K. Claessen and J. Hughes, "QuickCheck: A Lightweight Tool for Random Testing of Haskell Programs," in *Proceedings of the 2000 ACM SIGPLAN International Conference on Functional Programming (ICFP)*, ACM, 2000, pp. 268–279. doi: 10.1145/351240.351266.

[25] National Institute of Standards and Technology, "Security and Privacy Controls for Information Systems and Organizations," technical report NIST Special Publication 800-53 Revision 5, 2020. doi: 10.6028/NIST.SP.800-53r5.

# Appendix A: GraphFlow Language Example

The following is a GFL example demonstrating formal verification with mathematical properties:

```
# GraphFlow Language (GFL) Example - Minimal Non-Trivial Verified Workflow
# Demonstrates verification, composition, and assumptions

diagram [blueprint] :calculate-sum-of-squares-bounded "Calculate Sum of Squares (Bounded)":
  description: |
    Calculates the sum of squares for two values: a² + b².
    Demonstrates formal verification with contracts, branching,
    and an explicit assumption boundary.

  swimlanes:
    - "System":
        width: 2

  inputs: {
    .a: :number
    .b: :number
  }

  outputs: {
    .sum: :number
  }

  # Preconditions
  requires:
    - (:ne $.a null)
    - (:ne $.b null)

  # Postconditions
  ensures:
    - (:gte $.return.sum 0)

  properties:
    - (:is-deterministic $.return.sum)
    - (:is-total)
    - (:is-commutative $.a $.b)

  variables:
    $.aSquared: 0
    $.bSquared: 0

  model:
    1. [task] "Square a" @system --> 3:
      action:
        calls: (:multiply {
          .a: $.a
          .b: $.a
        })
        assigns: $.aSquared
      ensures: (:gte $.aSquared 0)

    2. [task] "Square b" @system --> 3:
      action:
        calls: (:multiply {
          .a: $.b
          .b: $.b
        })
        assigns: $.bSquared
      ensures: (:gte $.bSquared 0)

    3. [milestone] "Sum Squares" @system --> 4:
      requires: (:and (:ne $.aSquared null) (:ne $.bSquared null))
      action:
        calls: (:add {
          .a: $.aSquared
          .b: $.bSquared
        })
        assigns: $.sum
      ensures: (:gte $.sum 0)

    4. [decision] "Is sum within bound?" @system :yes--> 5a :no--> 5b:
      action:
        calls: (:condition {
          .yes: (:lte $.sum 1000)
        })
```

```
5a. [milestone] "Return Result" @system:
  action:
    calls: (:return {
      .sum: $.sum
    })
  ensures: (:lte $.return.sum 1000)

5b. [milestone] "Reject Unbounded Sum" @system:
  action:
    calls: (:throw "Sum exceeds allowed bound")
  # Explicit assumption boundary:
  properties:
    - (:assumed-boundary)
```

## Appendix B: Cognitive Testing Care Pathway

This appendix presents the Cognitive Testing Care Pathway as a GFL reconstruction of the workflow deployed in the 2025 pilot (see “Empirical Evaluation”).

```
# GraphFlow Language (GFL) - Cognitive Testing Care Pathway
#
# GFL reconstruction of the 2025 pilot workflow, demonstrating GraphFlow's three pillars:
# cohort search, workflow automation, and operational dashboards.

# Pillar I - Cohort Search
trigger "Cognitive Testing for Eligible Patients":
  trigger-type: :diagram
  description: "Enroll the eligible patients into the Cognitive Testing Care Pathway"
  active: true
  auto-start: true
  schedule:
    interval: :daily
  source:
    query: :cognitive-screening-eligible
  repeat:
    interval: :yearly
  calls: :cognitive-testing-care-pathway
  assignment:
    - (:assign-swimlane-contact {
        .swimlane: :patient
        .contactId: $.id
      })
    - (:assign-swimlane-contact-by-ext-id {
        .swimlane: :provider
        .extId: $.extData.usualProviderId
      })

query "Cognitive Screening Eligible":
  description: "Look for patients with upcoming appointments eligible for a cognitive screening"
  resource-type: :contact
  ext-type: "Patient"
  filters:
    - with: :upcoming-appointment
    - with: :over-60
    - without: :cognitive-screening-completed

# Pillar II - Formal Automation
diagram "Cognitive Testing Care Pathway":
  description: "Test patients for cognitive impairment"
  swimlanes:
    - "Patient"
    - "Staff"
    - "Provider"
    - "MedFlow"
    - "EHR"
  model:
    1. [task] "Order Cognitive Screening" @medflow --> 2:
      assigned: [:staff, :ehr]
      action:
        calls: (:create-lab-order {
          .orderType: "OTHER"
          .labOrder: "Cognitive Screening"
          .patientId: $.swimlanes.patient.contact.ext-id
          .orderingProviderId: $.swimlanes.provider.contact.ext-id
        })
        assigns: $.order
      ensures: (:ne $.order null)
    2. [task] "Complete Cognitive Screening" @patient --> 3:
      description: "Patient is able to complete this at home before their appointment"
      action:
        calls: (:send-text {
          .template: "cognitive-screening"
          .orderId: $.order.id
          .contact: $.swimlanes.patient.contact
        })
    3. [wait] "Cognitive Screening Result" @medflow --> 4:
      assigned: [:ehr]
      action:
        calls: (:await-with-tag {
          .resource: $.swimlanes.patient.contact
          .filters:
            - with: :cognitive-screening-completed
        })
    4. [decision] "Further testing recommended?" @provider :yes --> 5:
      action:
```

```
        calls: (:condition {
          .yes: (:with-tag $.swimlanes.patient.contact :cognitive-screening-positive)
        })
    5. [task] "Order Cognitive Assessment" @medflow --> 6:
      assigned: [:staff, :ehr]
      action:
        calls: (:create-lab-order {
          .orderType: "OTHER"
          .labOrder: "Cognitive Assessment"
          .patientId: $.swimlanes.patient.contact.ext-id
          .orderingProviderId: $.swimlanes.provider.contact.ext-id
        })
    6. [meeting] "Proctor Cognitive Assessment" @staff --> 7:
      assigned: [:patient]
      action:
        calls: :next
    7. [wait] "Cognitive Assessment Result" @medflow --> 8:
      assigned: [:staff, :ehr]
      action:
        calls: (:await-with-tag {
          .resource: $.swimlanes.patient.contact
          .filters:
            - with: :cognitive-assessment-completed
        })
    8. [decision] "Care Plan recommended?" @provider :yes --> 9:
      action:
        calls: (:condition {
          .yes: (:with-tag $.swimlanes.patient.contact :cognitive-assessment-positive)
        })
    9. [task] "Order Cognitive Care Plan" @medflow:
      action:
        calls: (:create-lab-order {
          .orderType: "OTHER"
          .labOrder: "Cognitive Care Plan"
          .patientId: $.swimlanes.patient.contact.ext-id
          .orderingProviderId: $.swimlanes.provider.contact.ext-id
        })

# Pillar III - Operational Dashboards
query "Cognitive Screening Ordered":
  resource-type: :contact
  ext-type: "Patient"
  filters:
    - with: :cognitive-screening-ordered

query "Cognitive Screening Completed":
  resource-type: :contact
  ext-type: "Patient"
  filters:
    - with: :cognitive-screening-completed

metric "Cognitive Screening Ordered":
  description: "Daily count of how many screenings have been ordered"
  query: :cognitive-screening-ordered
  aggregation: :count
  schedule:
    interval: :daily

metric "Cognitive Screening Completed":
  description: "Daily count of how many screenings are completed"
  query: :cognitive-screening-completed
  aggregation: :count
  schedule:
    interval: :daily
```

## Appendix C: Visual Workflow Example

This appendix provides a visual representation of the Cognitive Testing Care Pathway deployed in the 2025 pilot (see "Empirical Evaluation").

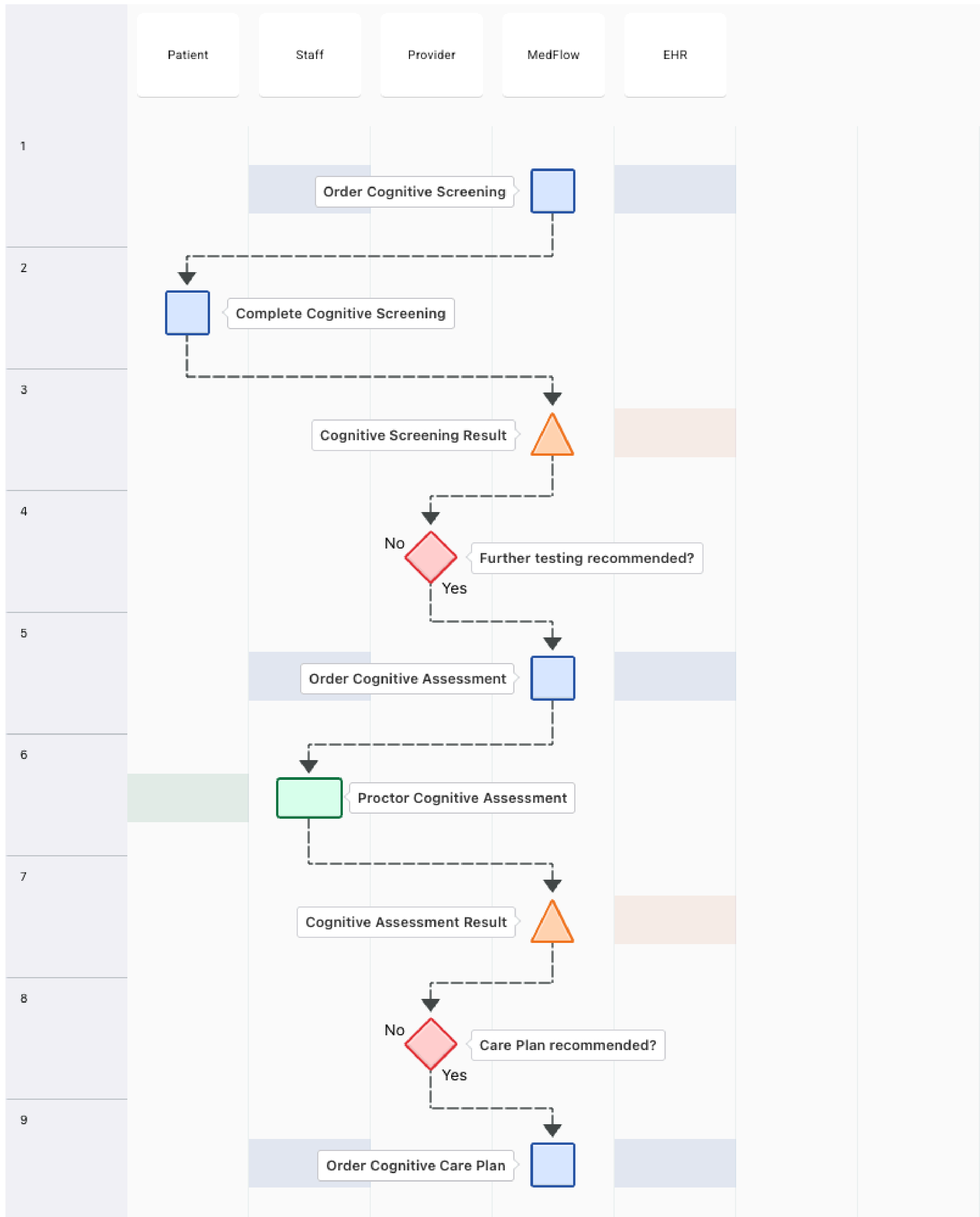

Figure 1: Cognitive Testing Care Pathway with swimlanes for Patient, Staff, Provider, MedFlow, and EHR.

# Appendix D: Formal Semantics and Theorem Statements

This appendix defines the formal semantics needed to state and interpret GraphFlow's correctness guarantees. All claims in this paper are relative to these semantics and the explicit assumptions stated herein.

GraphFlow uses two distinct but related semantics:

- **Compile-time (code generation)**: a restricted class of diagrams that are admissible to the compiler and subject to static verification.
- **Runtime (durable execution)**: an event-log-based execution model that supports retries, replay, and long-running workflows.

### D.1 Formal State

Workflow execution state is modeled as a finite mapping from variables to values:

- $\sigma : \mathrm{Var} \to \mathrm{Val}$ represents the workflow state, including inputs, intermediate variables, and outputs.
- Milestone or return primitives write the final result to $\sigma_{\mathrm{return}}$.

For durable execution, runtime state additionally includes an event log:

- $\sigma_R = (\sigma, \ell)$ where $\ell$ is an append-only sequence of events $e$.

### D.2 Core Artifacts

Let the following sets be defined:

- $\mathbb{V}$: node catalog (standalone nodes).
- $D \in \mathbb{D}$: diagrams.
- $I \in \mathbb{I}$: integrations.
- $A \in \mathbb{A}$: workflow automations.
- $\mathbb{X} = \mathbb{D} \cup \mathbb{I} \cup \mathbb{A}$: executable artifacts in the GraphFlow runtime.
- $P \in \mathbb{P}$: contacts (persons).

### D.3 Diagram Structure

A diagram is defined as a directed graph with associated metadata:

- **Graph**: $G = (V, E)$ where:
  - $V \subseteq \mathbb{V}$ is a finite set of diagram nodes.
  - $n \in V$ denotes a node in a diagram.
  - $n \in \mathbb{V}$ denotes a catalog (standalone) node.
  - $E \subseteq V \times \Sigma_E \times V$ is a labeled edge relation.
  - $\Sigma_E = \{\mathrm{to}, \mathrm{yes}, \mathrm{no}, \mathrm{maybe}\}$
- **Diagram**: $D = (G, \mu)$ where $\mu$ is a tuple of assignment functions:
  - $\lambda : V \to L$ assigns each node to a swimlane (lane) owner.
    - $L_c \subseteq L$ designates verifier-eligible (core) lanes.
    - $\lambda_{\mathrm{assigned}} : V \rightharpoonup 2^L$ optionally assigns additional swimlanes to a node.
  - $\tau : V \to \Sigma_V$ assigns a node type.
    - $\Sigma_V = \{\mathrm{task}, \mathrm{meeting}, \mathrm{report}, \mathrm{object}, \mathrm{decision}, \mathrm{queue}, \mathrm{wait}, \mathrm{milestone}, \mathrm{diagram}\}$
  - $\pi : V \to (\mathbb{R} \times \mathbb{R})$ assigns layout coordinates.
  - $\alpha : V \rightharpoonup \mathbb{A}$ optionally associates nodes with automation actions.
  - $w : V \rightharpoonup (\mathbb{R} \times \mathbb{R})$ optionally assigns cost/time weights.
  - $\delta : V \rightharpoonup \mathbb{D}$ optionally assigns a subdiagram.
  - $\psi : L \rightharpoonup \mathbb{P}$ optionally assigns a contact to a swimlane.

Let $\mathbb{D}$ be the set of all well-formed diagrams. Let $\mathbb{D}_c \subseteq \mathbb{D}$ be the subset of diagrams admissible to the compiler (acyclic and free of forked parallelism).

### D.4 Compile-time Semantics (big-step)

Compilation maps admissible diagrams to executable artifacts:

$\mathcal{C} : \mathbb{D}_c \to \mathbb{A} \cup \mathbb{I}$

Let $X = \mathcal{C}(D)$ for $D \in \mathbb{D}_c$.

Diagram execution at compile-time is defined by a big-step evaluation relation:

$\langle D, \sigma \rangle \Rightarrow \sigma'$

meaning that executing diagram $D$ from initial state $\sigma$ yields final state $\sigma'$.

Compiled artifacts execute according to their source diagram:

$\langle X, \sigma \rangle \Rightarrow \sigma' \Longleftrightarrow \langle D, \sigma \rangle \Rightarrow \sigma'$

Composition obligations are generated along edges $(u, s, v) \in E$ where $s \in \Sigma_E$: for any edge from node $u$ to node $v$, the postcondition of $u$ must imply the precondition of $v$. Formally, for all edges $(u, s, v) \in E$ and all states $\sigma'$:

$\forall \sigma' : \mathrm{Ensures}_{u(\sigma')} \Rightarrow \mathrm{Requires}_{v(\sigma')}$

This ensures that sequential composition preserves correctness assumptions across node boundaries.

### D.5 Runtime Semantics (small-step)

Runtime execution is modeled as a small-step transition relation that emits events:

$\langle n, (\sigma, \ell) \rangle \to (\sigma', \ell + e)$

Nodes assigned to verifier-ineligible lanes represent effect boundaries. Formally:

- If $\lambda(n) \notin L_c$, then execution of $n$ obtains its outcome from an external source (external system, AI, or human judgment).
- The observed outcome MUST be recorded as an event $e$ appended to the event log $\ell$.
- Replay consults $\ell$ rather than re-evaluating boundary effects.

This requirement ensures deterministic replay for a fixed initial state and event log.

### D.6 Contracts

Correctness is expressed relative to explicit contracts.

Each node or diagram may declare predicates:

- $\text{Requires}_{n(\sigma)}$ — precondition.
- $\text{Ensures}_{n(\sigma')}$ — postcondition.

Predicates are deterministic expressions over workflow state and are interpreted only for nodes whose semantics are defined in the compile-time model. Boundary nodes are not proven correct. Their outcomes are treated as recorded assumptions.

### D.7 Theorems

The following theorems provide the formal basis for GraphFlow's correctness claims.

**Theorem 1 (Contract soundness for compiled artifacts: partial correctness).** Assume:

- A compiled artifact $X$ is produced from a diagram $D$ with $X = \mathcal{C}(D)$.
- All referenced actions are deterministic and satisfy their declared contracts.
- All generated proof obligations for $X$ are accepted by the verifier.

Then, for all workflow states $\sigma$:

If $\text{Requires}_{X(\sigma)}$ holds and $\langle X, \sigma \rangle \Rightarrow \sigma'$, then $\text{Ensures}_{X(\sigma')}$ holds.

This establishes partial correctness. Termination requires separate assumptions or proofs.

**Theorem 2 (Deterministic replay: runtime).** Assume:

- All nondeterministic outcomes influencing control flow are recorded in the event log $\ell$.
- Replay reads outcomes from $\ell$ instead of consulting external sources.

Then replay is deterministic for a fixed initial state $\sigma_0$ and event log $\ell$.